%% file: main.tex
\documentclass{article}
\usepackage{amsmath}
\usepackage{graphicx} 
\usepackage{hyperref} 
\usepackage{indentfirst}
\usepackage{multicol} 
\usepackage{float}  
\usepackage{multirow}
\usepackage{amssymb}
\usepackage{algpseudocode}

\usepackage{authblk}
\usepackage[ruled,vlined]{algorithm2e}
\usepackage[backend=biber, style=numeric]{biblatex} 
\usepackage[top=1in, bottom=1in, left=0.8in, right=0.8in]{geometry}
\title{TSDW: A Tri-Stream Dynamic Weight Network for Cloth-Changing Person Re-Identification}

\author[1]{Ruiqi He\thanks{8008122083@email.ncu.edu.cn}}
\author[1]{Zihan Wang\thanks{8008122070@email.ncu.edu.cn}}
\author[1,*]{Xiang Zhou\thanks{zhouxiang2021@tongji.edu.cn}}
\affil[1]{School of Software, Nanchang University, Nanchang, 330047, China}
\affil[*]{Department of Computer Science and Technology, Tongji University, Shanghai,201804, China}
\date{February 2025}

\begin{document}

\maketitle
\begin{abstract}
Cloth-Changing Person Re-identification (CC-ReID) aims to solve the challenge of identifying individuals across different temporal-spatial scenarios, viewpoints, and clothing variations. This field is gaining increasing attention in big data research and public security domains. Existing ReID research primarily relies on face recognition, gait semantic recognition, and clothing-irrelevant feature identification, which perform relatively well in scenarios with high-quality clothing change videos and images. However, these approaches depend on either single features or simple combinations of multiple features, making further performance improvements difficult. Additionally, limitations such as missing facial information, challenges in gait extraction, and inconsistent camera parameters restrict the broader application of CC-ReID. To address the above limitations, we innovatively propose a Tri-Stream Dynamic Weight Network (TSDW) that requires only images. This dynamic weighting network consists of three parallel feature streams: facial features, head-limb features, and global features. Each stream specializes in extracting its designated features, after which a gating network dynamically fuses confidence levels. The three parallel feature streams enhance recognition performance and reduce the impact of any single feature failure, thereby improving model robustness. Extensive experiments on benchmark datasets (e.g., PRCC\cite{Yang2021PersonRB}, Celeb-reID\cite{topdropblock2020}, VC-Clothes\cite{wan2020person}) demonstrate that our method significantly outperforms existing state-of-the-art approaches.

\noindent{\textbf{keywords:}Cloth-changing person re-identification, Person re-identification,Convolutional neural networks,Tri-Stream Dynamic Weight Network.}
\end{abstract}

\begin{multicols}{2} 
\section{INTRODUCTION}

Person Re-Identification (ReID) aims to match the same individual from different perspectives and plays an important role in applications within the public safety domain. In recent years, with the development of deep learning technologies and the increasing societal emphasis on public safety, traditional person re-identification methods have struggled to address more complex societal challenges, such as varying temporal and spatial scenes, different clothing features, and inconsistencies in data across different camera sources, all of which call for more advanced research topics.

As a result, there has been a growing focus on Clothing Change Person Re-Identification (CC-ReID), where the core task is to capture the same individual wearing different clothing from various data sources. Existing CC-ReID methods can be categorized into several types. Gu et al. \cite{Gu_2022} utilized Clothes-based Adversarial Loss to discover clothing-invariant features from images. Wang et al. \cite{Wang_2023} embedded 2D human images into 3D human models. Bansal et al. \cite{bansal2022cloth} proposed a Vision Transformer framework to capture abstract gait information. Wu et al. \cite{wu2022identitysensitiveknowledgepropagationclothchanging} suggested using prior facial knowledge to reconstruct lost facial details in low-resolution images, enhancing the ability to capture invariant biometric features. However, due to the difficulty in ensuring that the data is complete and comprehensive, focusing solely on extracting one type of information from images may lead to significant losses. For instance, in facial feature recognition, the individual may be facing away from the camera; in the case of limb information extraction, the individual in the image may be missing a limb or even facial information. Undoubtedly, this poses challenges for methods that focus on extracting a single feature. Other researchers have aimed to decouple clothing-invariant information from images, but studies such as those by CHAN et al. \cite{10.1145/3584359} indicate that the ability to decouple features diminishes when certain input data is affected by additional complex environmental factors (e.g., differences in camera parameters, occlusion, strong lighting conditions).

\begin{figure}[H]
    \centering
    \includegraphics[width=0.8\columnwidth]{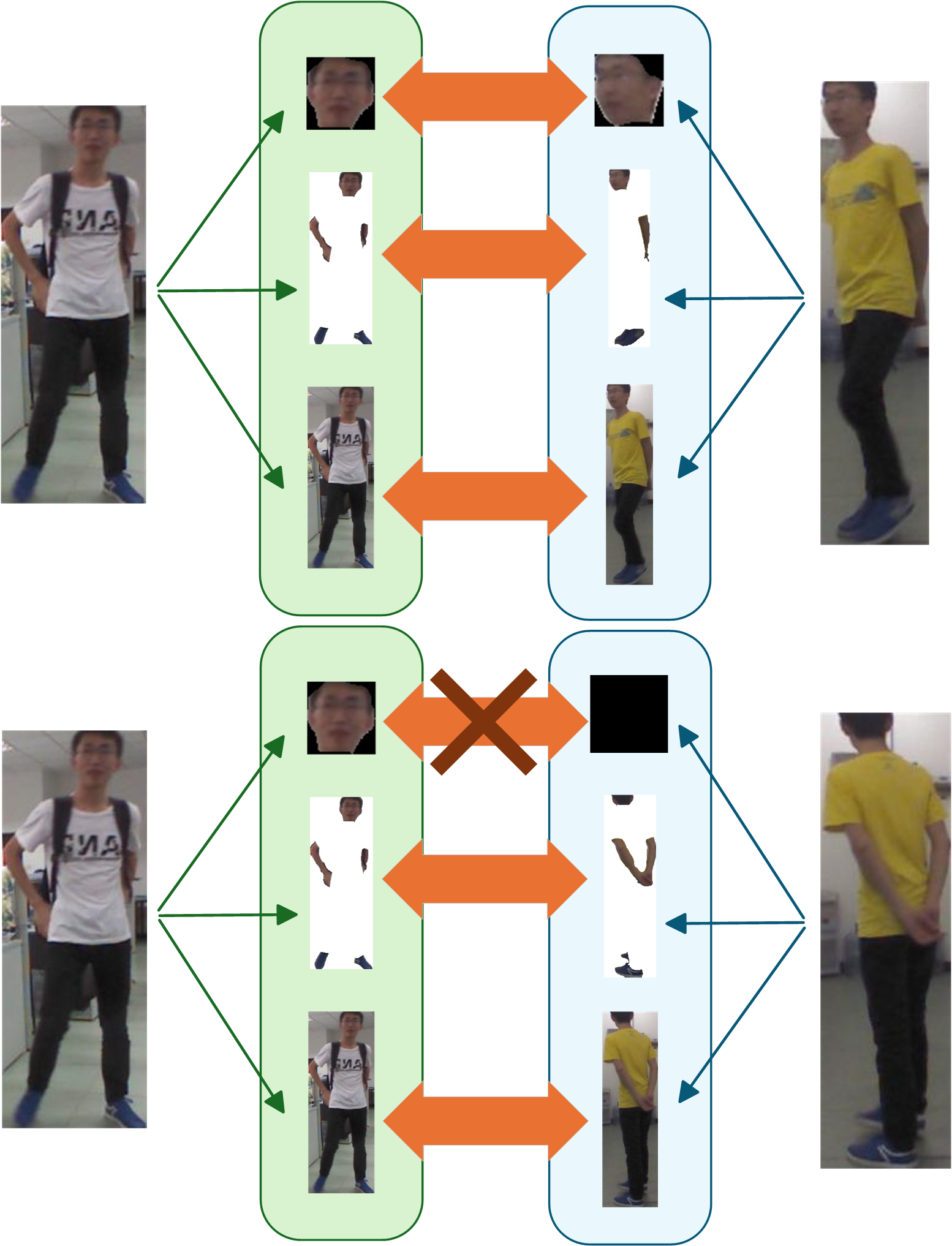}
    \caption{\textbf{Comparison of the face, head-limb, and global image.} From these three different perspectives, we can simultaneously compare and identify whether they belong to the same person.However, it's important to note that facial images are not always available across all perspectives. For example, in the first comparison, we can proceed with all three angles, while in the second comparison, we are limited to only two angles.}
    \label{fig:三流对比小图}
\end{figure}

To address the diverse challenges of real-world situations and improve model robustness, this paper innovatively proposes a Tri-Stream Dynamic Weight Network (TSDW). The model primarily consists of a Semantic Human Parsing (SCHP~\cite{li2020self}) module, parallel feature streams, a dynamic fusion mechanism, and a conditional gating strategy. First, the Semantic Human Parsing (SCHP) module generates three region-specific inputs: facial crops, limb-focused masks, and the global image. Next, each input is processed by dedicated sub-networks: a facial stream with zero-vector masking capability, a head-limb attention stream, and a global stream that is clothing-invariant. By obtaining the final result through weighted aggregation, we propose a dynamic weighted decision module that intelligently fuses the features of all streams using a hierarchical gating mechanism with learnable weights, based on facial detection confidence and clothing interference levels, to enhance accuracy and improve robustness. The proposed method outperforms all state-of-the-art methods on benchmark datasets (e.g., PRCC, Celeb-reID, VC-Clothes).

The contributions of this paper are as follows:

\begin{itemize}
    \item We propose a novel Tri-Stream Dynamic Weight Network (TSDW) that processes facial, head-limb, and global streams in parallel, thereby avoiding errors caused by the failure of a single feature.
    \item We employ a hierarchical gating mechanism with learnable weights, intelligently fusing the features of all streams through three-way decision, thereby improving accuracy and enhancing robustness.
    \item Experimental results on the PRCC, Celeb-reID, and VC-Clothes datasets demonstrate that our model outperforms existing models across multiple capabilities.
\end{itemize}


\section{RELATED WORKS}

\subsection{Person Re-Identification (ReID)}

Person Re-Identification (ReID) has a rich history of practical applications in scenarios where clothing remains unchanged. This includes several major ReID methods such as facial recognition \cite{dietlmeier2020importantfacespersonreidentification}, gait feature extraction \cite{zhang2023realgaitgaitrecognitionperson}, and semantic extraction of invariant biometric features \cite{li2023clothesinvariantfeaturelearningcausal}. These methods have proven to be quite reliable in standard ReID tasks; however, under long-term temporal and spatial conditions, individuals typically change their clothing, making it unrealistic to expect the target individual to remain in the same attire. Consequently, traditional ReID methods may lose their practical value in general scene applications.

\subsection{Clothing Change Person Re-Identification (CC-ReID)}

To address the challenges posed by clothing changes in general scenarios, researchers have focused on extracting clothing-invariant features to identify the unchanging characteristics within images, leading to the development of Clothing Change Person Re-Identification (CC-ReID). Through the exploration of invariant feature information, researchers have proposed several focal areas, including body information extraction and clothing feature decoupling.

Body information extraction encompasses gait feature recognition methods \cite{7899654}\cite{10339887}\cite{jin2022clothchangingpersonreidentificationsingle} and part information extraction, which is a widely used approach that mitigates the negative impacts of clothing changes by focusing on identifying key points of individuals in images, thereby reconstructing a 3D human skeleton and matching it against all gait candidates to find the corresponding individual. However, the interference of clothing on skeleton re-identification is difficult to eliminate, making it challenging to further improve accuracy. Additionally, focusing solely on gait information may lead to the neglect of other clothing-invariant features that still hold recognition value, such as facial and body shape information. Furthermore, part information extraction primarily concentrates on invariant features brought by specific body parts, such as the most accurate facial recognition, exposed limb shape recognition, and human contour information extraction \cite{qian2020longtermclothchangingpersonreidentification}. As mentioned earlier, while facial recognition and other methods for extracting exposed biometric information are effective when image quality is sufficiently high, they face challenges when cross-source image quality declines or when strong lighting, clothing, or other factors obscure and interfere with body parts\cite{10.1145/3584359}. Body information reconstruction aims to recover relevant information from 2D images; for instance, when facial information is compromised, prior knowledge can be used to reconstruct the face, thereby obtaining high-quality facial data \cite{wu2022identitysensitiveknowledgepropagationclothchanging}. Additionally, 2D images of individuals can be projected onto 3D human models \cite{Wang_2023} to reconstruct complete human information. Reconstruction methods can effectively resist damage caused by information loss; however, they incur the highest computational costs in terms of pre-trained learning and inference among all methods, making them less satisfactory in time-sensitive and cost-sensitive tasks. In summary, methods relying on specific body information are only effective under certain conditions and do not comprehensively consider how to classify decisions across different images or how to select the most effective body information in various scenarios.

Clothing feature decoupling has led to the emergence of methods aimed at extracting high-dimensional feature information that is unaffected by clothing changes, including clothing masking \cite{qian2020longtermclothchangingpersonreidentification}, GAN adversarial networks \cite{10.1145/3584359}, causal intervention simulation \cite{GoodisBad}, multi-positive classification search \cite{Gu_2022}, and clothing information stripping \cite{wei2025multipleinformationpromptlearning}. These methods require models to focus on clothing-invariant information, achieving a more comprehensive utilization of image feature information. However, most methods operate with a single feature attention module, which still lacks completeness in the overall decision-making process.

Therefore, in our work, we will focus on exploring the comprehensive utilization of diverse identity feature information, decision-making for low-quality feature information, and enhancing model robustness.

\section{METHODOLOGY}

\begin{figure*}
    \centering
    \includegraphics[width=1\linewidth]{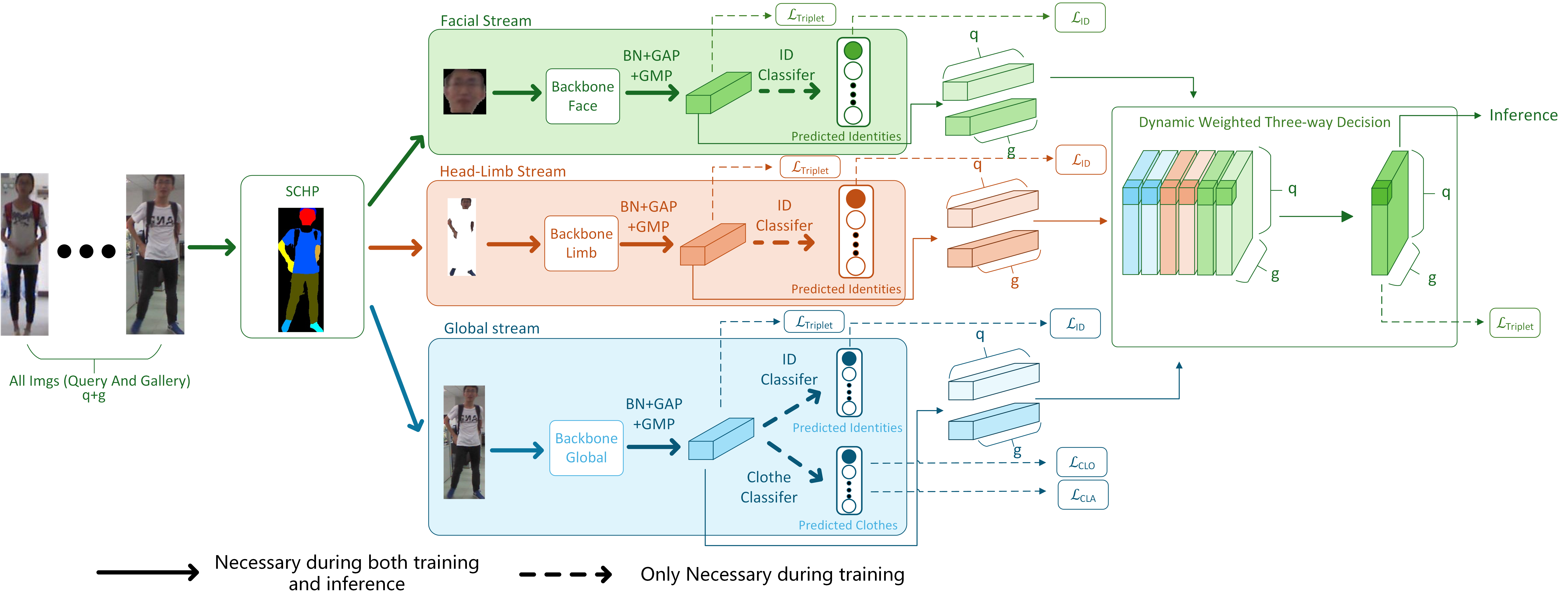}
    \caption{The proposed TSDW framework consists of an SCHP preprocessing module, three complementary feature extraction streams, and a dynamic weighted three-way decision module. The three parallel streams adopt different strategies to extract feature representations. The dynamic weighted three-way decision module adaptively assigns weights to each feature stream based on the input query and gallery features, ultimately generating a q×g similarity matrix, where q and g represent the number of images in the query and gallery sets, respectively. This architecture effectively enhances person re-identification matching performance in complex scenarios through its dynamic weighting mechanism.}
    \label{fig:TSDW}
\end{figure*}

To address the challenge of clothing-changing person re-identification, we propose a novel Tri-Stream Dynamic Weight Network (TSDW) that adaptively integrates facial features, limb features, and clothing-independent global features. As shown in Figure \ref{fig:TSDW}, the framework consists of three main components: 1) an SCHP preprocessing module for region segmentation; 2) three parallel feature streams, each extracting different aspects of features; and 3) a dynamic fusion mechanism using sequential three-way decision.

The model operates through three progressive stages: First, the Semantic Human Parsing (SCHP) module generates three region-specific inputs—facial crops, head-limb images, and global images. Second, each input is processed by specialized subnetworks: a facial stream with zero-vector masking capability, a head-limb attention stream, and a global stream focusing on clothing-independent features. Inspired by the Mixture of Experts (MOE) paradigm, we can view the three streams as three different expert models whose weighted outputs determine the final result. Therefore, we propose a dynamic weighting three-way decision module that dynamically fuses features from all streams through a hierarchical gating mechanism with learnable weights, based on facial detection confidence and clothing interference levels, to achieve optimal performance.
\subsection{SCHP Pre-processing}

To enable the three streams to focus on extracting features from different regions, we use the Semantic Human Parsing (SCHP) module for preprocessing the input image. This module divides the image into different parts based on various features, allowing each stream to focus on facial, head-limb, and global information. This not only enhances the robustness of the model but also improves its performance.

We chose a two-step approach to process the image: first, the SCHP module generates semantic segmentation masks to represent the positions of different body parts; then, these masks are used for cropping or processing, resulting in three distinct images as inputs for the subsequent streams:

\textbf{Facial Image:} We retain only the facial region. By cropping the face, the rest of the image is turned into a black background mask. A minimum facial area threshold, $\theta_{\text{face}}$, is set, and if no face is detected or the face is smaller than this threshold, the module outputs a $1 \times 1$ black image to indicate the absence of facial features. The cropped facial image contains the facial information features extracted by the model.

The mathematical expression for the facial image preprocessing flow is given in equation (1), where $S$ represents the $SCHP$ segmentation function, outputting the mask, and $Crop$ is the cropping operation:

\[
\begin{cases}
I_{\text{face}} = \operatorname{Crop}(I, S(I)_{\text{face}}) & \text{if } \| S(I)_{\text{face}} \|_1 > \theta_{\text{face}} \\
I_{\text{face}} = \mathbf{0}_{1 \times 1} & \text{otherwise}
\end{cases}
\tag{1}
\]

\textbf{Head-Limb Image:} We retain the head (including face and hair) and limb regions (such as arms, legs, and feet or shoes), while the remaining areas are filled with a white mask. This ensures that the head-limb stream can only focus on body parts, completely excluding clothing features. Although shoes are technically clothing items, they are preserved in this approach because they often display distinctive characteristics that help establish a person's identity, thereby improving the model's performance.

The mathematical expression for the head-limb image preprocessing flow is given in equation (2), where $WhiteFill$ represents filling the non-target areas with white:
\[
I_{\text{head-limb}} = \text{WhiteFill}(I, S(I)_{\text{head} \cup \text{limb}})
\tag{2}
\]

\textbf{Global Image:} This image retains all the global information without emphasizing any specific body part or feature. The global image stream extracts global information features, balancing body and clothing information along with other features.
\[
I_{\text{global}} = I
\tag{3}
\]

\subsection{Three Types of Feature Stream}
\subsubsection{Facial Stream}

The facial stream input consists of cropped facial images. Unlike other features, facial features have unique advantages in person re-identification tasks, achieving higher accuracy. Especially in clothing-change scenarios, facial information remains unaffected by clothing variations, thus providing the system with more robust discriminative evidence. However, faces also present uncertainty, as not all images contain facial features, so we need to address this uncertainty in subsequent processing.

We employ ResNet-50 as the backbone network for the facial stream, inputting previously processed facial images and combining label-smoothed cross entropy loss\cite{szegedy2015rethinkinginceptionarchitecturecomputer} with hard triplet loss\cite{hermans2017defensetripletlossperson} during training to enhance facial feature extraction capabilities. We use cosine distance as the similarity metric. To properly handle images without facial features, when the input is a completely black image (previously agreed to indicate absence of facial features), the network outputs a zero vector, ensuring consistent batch sizes and enabling normal processing of cases where facial features are absent.
\[
f_{face} = \begin{cases} 
\vec{0} & \text{if } I_{face} \text{ is a black image} \\
F_{ResNet}(I_{face}) & \text{otherwise}
\end{cases}
\tag{4}
\]
\begin{figure}[H]
    \centering
    \includegraphics[width=0.8\columnwidth]{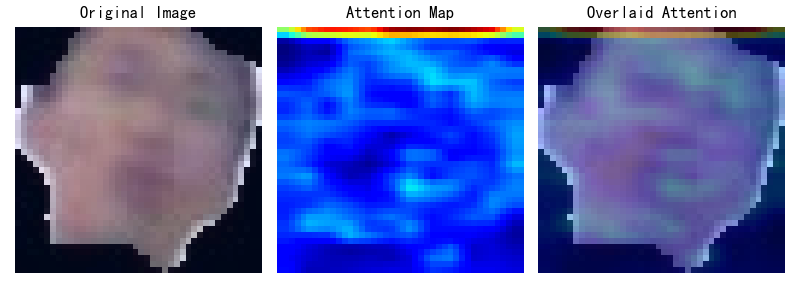}
    \caption{Attention Heat Map of Facial Stream}
    \label{fig:Attention Heat Map of Facial Stream}
\end{figure}

\subsubsection{Head-Limb Stream}

Due to the unique characteristics of facial features, we need normalized features that aren't affected by clothing changes, which is precisely what head and limb features provide. Unlike facial information, the exposed parts of the head and limbs are relatively resistant to clothing variations, and these exposed body parts can provide stable identity information without the risk of being absent like facial features often are. Therefore, the head-limb stream can provide additional feature information for pedestrian re-identification that is independent of clothing changes, and when facial features are missing, the stable characteristics of limbs ensure the model can maintain high accuracy even in cross-clothing scenarios.

In this stream, we use ResNet-50 as the backbone network for our head-limb stream, with input images of heads and limbs preprocessed through SCHP. The network is trained using both label-smoothed cross entropy loss and hard triplet loss.We use cosine distance as the similarity metric.
\begin{figure}[H]
    \centering
    \includegraphics[width=0.8\columnwidth]{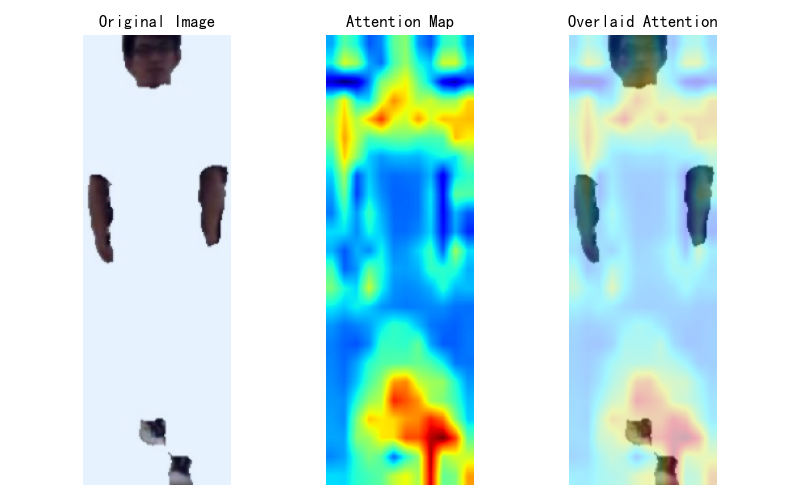}
    \caption{Attention Heat Map of Head-Limb Stream}
    \label{fig:Attention Heat Map of Head-Limb Stream}
\end{figure}

\subsubsection{Global stream}

According to research by Wang et al.\cite{wang2024learning} and Gu et al.\cite{Gu_2022}, excessive stripping of clothing features actually leads to a decrease in recognition capability. Therefore, in addition to the facial flow and head-limb flow, we need to design a flow that incorporates certain clothing features to prevent the architecture from excessively eliminating clothing characteristics, thereby enhancing the model's robustness.

In the global stream, we aim to utilize the complete RGB image by extracting features that contain the overall appearance of both the person and their clothing, thus balancing clothing and identity features. Unlike other flows (such as facial flow and head-limb flow) that focus on specific information, the global stream concentrates on the holistic information of the image, including clothing, body shape, posture, and accessories. Particularly in clothes-changing scenarios, the main challenge for the global stream is how to maintain robust recognition despite clothing variations.

To address this challenge, we draw inspiration from the optimization method proposed by Gu et al.\cite{Gu_2022} to improve RGB-based pedestrian re-identification performance. we use ResNet-50 as the backbone network for our global stream. First, the model is trained for a certain number of epochs using cross-entropy loss and hard triplet loss to enable stable person identification. Then, a clothing classifier is trained by minimizing the clothing classification loss, thereby encouraging the classifier to learn clothing-related features. After updating the parameters of the clothing classifier, the network backbone is optimized through the joint application of cross-entropy loss, hard triplet loss, and Clothes-based Adversarial Loss CAL $\mathcal{L}_{CA}$, enabling the global stream not only to perform identity recognition but also to maximally avoid over-reliance on clothing variations. This approach allows the global stream to learn a balance between person and clothing information, providing mixed information about both the individual and their attire for subsequent fusion.
\[
\mathcal{L}_{CA} = -\sum_{i=1}^{N}\sum_{c=1}^{N_C} q(c) \log \frac{e^{(f_i \cdot \varphi_c / \tau)}}{e^{(f_i \cdot \varphi_c / \tau)} + \sum_{j \in S_i^-} e^{(f_i \cdot \varphi_j / \tau)}},
\tag{5}
\]

\[
q(c) = 
\begin{cases} 
\frac{1}{K}, & c \in S_i^+ \\
0, & c \in S_i^-
\end{cases},
\tag{6}
\]
\begin{figure}[H]
    \centering
    \includegraphics[width=0.8\columnwidth]{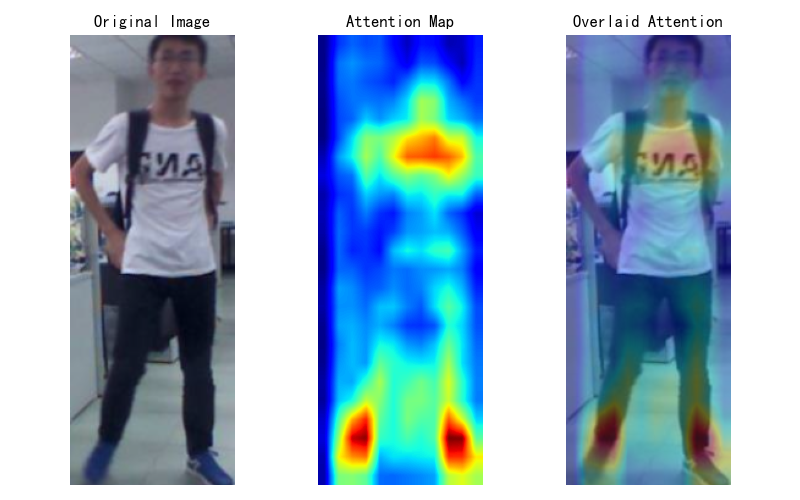}
    \caption{Attention Heat Map of Global Stream}
    \label{fig:Attention Heat Map of global stream}
\end{figure}

\subsection{Dynamic Weighted Three-way Decision(DWT)}

Due to the uncertainty of facial features, as they are not present in all images, we need to implement special processing logic for facial features and also need to integrate head-limb and global feature information. Therefore, we propose a dynamic weighted Three-way Decision module. Following the concept of Mixture of Experts (MOE\cite{Jacobs_Jordan_Nowlan_Hinton_1991}), we can view the three different types of features as three different expert models, each judging person ID from different perspectives. Directly fusing these three feature types might create difficulties when comparing an image with facial features to one without. Instead, we adopt a pairwise comparison approach with sequential three-way decision-making to output the final weights for feature distance fusion. This effectively mitigates the uncertainty caused by the absence of facial features.

When comparing queries with the gallery, first, since facial features have inherent uncertainty and might be the first features excluded from fusion, our first layer uses facial confidence for judgment, establishing a baseline for fusion. However, when facial features are unreliable or in an intermediate state, we need to incorporate head-limb features and global features for auxiliary judgment. Therefore, the second and third layers of the three-way decision process determine how to utilize head-limb and global features for assistance. Based on this overall approach, we can begin our design.

For this purpose, we first need multiple trainable confidence score MLP layers to output confidence scores as the basis for sequential three-way decision-making, as well as gating networks to continue training for weights. Therefore, we design three confidence networks and four gating networks to output different weights, informing us which weights to use for fusion.All threshold parameters ($\alpha$ and $\beta$ with their respective subscripts) mentioned in the following decision layers are trainable parameters that are optimized during the learning process, and their values range between 0 and 1.

\input{pseudocode}

\textbf{First Layer Decision (Based on Facial Confidence):}

Based on ${confidence_f}$, we output the facial confidence $C_{f}$, where $q_{f}$ and $g_{f}$ are the query facial features and gallery facial features respectively.If either $q_f$ or $g_f$ is a zero vector (indicating no face), the $C_{f}$ is 0.
\[
C_{f} =\sigma({confidence_f}(q_{f},g_{f}))
\tag{7}
\]

Based on the derived facial confidence, we decide whether to directly use facial features or proceed to the second or third layer decision
\[
\mathbf{w} =  \begin{cases} [1,0,0], & C_{f} > \alpha_{f} \\ \text{Second Layer Decision}, & \beta_{f} \leq C_{f} \leq \alpha_{f} \\ \text{Third Layer Decision}, & C_{f} < \beta_{f} \end{cases}
\tag{8}
\]

Where: 
\begin{itemize}
    \item $\sigma$ is the Sigmoid function, used to map confidence scores to the $[0,1]$ interval.
    \item $\alpha_f$ and $\beta_f$ are preset facial confidence thresholds, satisfying $\alpha_{f} > \beta_{f}$, representing the boundaries for complete trustworthiness and complete untrustworthiness respectively.
    \item The weight vector $\mathbf{w} = [w_f, w_l, w_g]$ corresponds to the fusion weights for facial, head-limb (head+body), and global features. When $C_{f} > \alpha_{f}$, only facial features are retained ($w_f=1$), with other weights being $0$
\end{itemize}

\textbf{Second Layer Decision (When Face is Available but Not Completely Reliable):}

Calculate the (head-limb)-global confidence $C_{l-g_1}$ for second-layer judgment
\[
C_{l-g_1} = \sigma({confidence_{{l-g}_1}}([q_{l};q_{g}],[g_{l};g_{g}]))
\tag{9}
\]

Since the second layer decision occurs when facial features are insufficient for complete judgment, we will fuse other features for assistance. Based on the derived (head-limb)-global confidence $C_{l-g_1}$, we decide whether to fuse the face with one of the other features or to fuse all three features

\[
\mathbf{W} =  \begin{cases}  
\text{Gating}_{f-g}(q_{f},q_{g},g_{f},g_{g}), & C_{l-g_1} > \alpha_{l-g_1} \\ \text{Gating}_{f-l}(q_{f},q_{l},g_{f},g_{l}), & C_{l-g_1} < \beta_{l-g_1} \\ 
\text{Gating}_{all}(q_{all},g_{all}), & \text{otherwise} \end{cases}
\tag{10}
\]
Where: 
\begin{itemize}
    \item $\sigma$ is the Sigmoid function, calculating the joint (head-limb)-global confidence $C_{l-g_1}$.
    \item $\alpha_{l-g_1}$ and $\beta_{l-g_1}$ are second-layer thresholds, satisfying $\alpha_{l-g_1} > \beta_{l-g_1}$, used to determine the dominance of limb or global features.
    \item Various Gating functions generate weights through attention mechanisms or multimodal interaction:
    \begin{itemize}
        \item $\text{Gating}_{\text{f-g}}$ outputs weights $[w_f, 0, w_g]$, constrained such that $w_f + w_g = 1$ (via Softmax).
        \item $\text{Gating}_{\text{f-l}}$ outputs weights $[w_f, w_l, 0]$, constrained such that $w_f + w_l = 1$.
        \item $\text{Gating}_{\text{all}}$ outputs all weights $[w_f, w_l, w_g]$, constrained such that $w_f + w_l + w_g = 1$.
    \end{itemize}
    \item Weight allocation strategy: If (head-limb)-global confidence is high ($C_{l-g_1} > \alpha_{l-g_1}$), preferentially fuse facial and global features; if confidence is low ($C_{l-g_1} < \beta_{l-g_1}$), fuse facial and limb features; otherwise, jointly fuse all three.
\end{itemize}

\textbf{Third Layer Decision (When Face is Not Available):}

Calculate the (head-limb)-global confidence $C_{l-g_2}$ for third-layer judgment:
\[
C_{l-g_2} = \sigma(confidence_{{l-g}_2}([q_{l};q_{g}],[g_{l};g_{g}]))
\tag{11}
\]
When facial features are unreliable, select a fusion strategy based on the confidence $C_{l-g_2}$:
\[
\mathbf{W} =  \begin{cases} 
[0,0,1], & C_{l-g_2} > \alpha_{l-g_2} \\
[0,1,0], & C_{l-g_2} < \beta_{l-g_2} \\
\text{Gating}_{l-g}(q_{l}, q_{g}, g_{l}, g_{g}), & \beta_{l-g_2} \leq C_{l-g_2} \leq \alpha_{l-g_2}
\end{cases}
\tag{12}
\]
Where:

\begin{itemize}
    \item $\text{Gating}_{l-g}$ outputs two weights $[0, w_l, w_g]$, constrained such that $w_l + w_g = 1$ via softmax
    \item $\alpha_{l-g_2}$ and $\beta_{l-g_2}$ are preset (head-limb)-global confidence thresholds, satisfying $\alpha_{l-g_2} > \beta_{l-g_2}$
\end{itemize}

\textbf{Dynamic Distance Fusion:}

The final distance matrix is calculated as:
\[
D_{final} = \sum_{k \in \{f,l,g\}} w_k^{(q,g)} \cdot d_k(q_k, g_k)
\tag{13}
\]
Where the cosine distance $d_k = 1 - \frac{q_k \cdot g_k}{\|q_k\| \|g_k\|}$

Finally, we output the mixed distance matrix, trained using a hard triplet loss to ensure accurate mixing:
\[
\mathcal{L}_{Triplet} = \max(d_{an}^{hard} - d_{ap}^{hard} + \mathbf{m}, 0)
\tag{14}
\]

Where:
\begin{itemize}
    \item $d_{an}^{hard}$ is the distance of the hardest negative sample pair,
    \item $d_{ap}^{hard}$ is the distance of the hardest positive sample pair,
    \item $\mathbf{m}$ is a predefined margin.
\end{itemize}

\section{EXPERIMENTS AND RESULTS}

\subsection{Implementation Details}
We employ a carefully designed training strategy, including a batch size of 32 (4 identities of pedestrians, 8 images per identity), with a unified image size of 384×192 pixels. First, the three-stream independent training uses the Adam optimizer (learning rate 0.00045, weight decay 5e-4), and the OneCycleLR learning rate strategy, which includes a warm-up phase covering 30\% of the total training and momentum adjustment (0.85-0.95) that inversely changes with the learning rate. The training lasts for 100 epochs, and the global stream introduces CAL loss after the 25th epoch to enhance feature learning. Then, the decision fusion training of the three streams is based on the previously pre-trained three-stream model, using the Adam optimizer (initial learning rate 5e-6, weight decay 5e-4) and the MultiStepLR learning rate strategy (decay factor of 0.1 at the 20th and 40th epochs). In the first 10 epochs, the parameters of the three streams are frozen, and after that, the entire model architecture is jointly optimized. The model is then trained for another 50 epochs to ensure that all components are fully learned and effectively fused, achieving the best performance.

\subsection{Datasets}
\textbf{Celeb-reID}\cite{topdropblock2020}

Celeb-reID uses street snap-shots of celebrities acquired from the Internet. There are 34,186 images with 1,052 IDs. The probability of clothing change for a person is 70\% on average. That is, both clothing-change case (70\%) and no-clothing-change case (30\%) coexist in the training and test sets of Celeb-reID.

\textbf{PRCC}\cite{Yang2021PersonRB}

The PRCC dataset, created by Yang's team, focuses on person re-identification research under clothing change conditions. This dataset includes 33,698 portrait images of 221 different individuals captured through three distinct cameras (A, B, C). On average, each subject has approximately 50 photos under each camera, totaling about 152 different images per person. Cameras A and B capture individuals wearing the same clothing but located in different rooms, while Camera C records the same individuals wearing different clothing and photographed on another day. Beyond clothing variations, this dataset also encompasses multidimensional factors such as lighting conditions, occlusion situations, postures, and angles. In the experimental setup, images of 150 individuals are used for model training, with the remaining 71 used for testing. During testing, one image per person from Camera A constructs the retrieval gallery, while all images from Camera B (same clothing) and Camera C (different clothing) serve as the query set.

\textbf{VC-Clothes}\cite{wan2020person}

VC-Clothes is a virtual character dataset generated using GTA5 gaming technology. This collection contains 512 independent identities, photographed in 4 different scenes, with an average of 9 images per identity in each scene, totaling 19,060 images. In our research, we divide the data equally by identity: half (256 identities) for algorithm training and the other half for performance evaluation. During the testing phase, we randomly extract 4 images of each identity from each camera position as query samples, with the remaining images forming the retrieval gallery. In the final configuration, the training data includes 9,449 images, while the test set consists of 1,020 query images and 8,591 gallery images.

\subsection{Evaluation Metrics}
Pedestrian re-identification systems are typically evaluated using two key metrics: Rank-1 accuracy and mean Average Precision (mAP). Rank-1 accuracy measures the percentage of correctly matched images ranked first in the gallery for a given query, directly reflecting the system’s ability to make accurate initial matches. Mean Average Precision (mAP) offers a more comprehensive evaluation by considering the quality of the entire ranking, averaging precision values at different recall levels across all queries. While Rank-1 accuracy focuses on the success rate of top matches, mAP provides an overall assessment that accounts for multiple instances of the same identity, making it especially valuable in complex scenarios such as re-identification under clothing variations. These two complementary metrics together offer a thorough evaluation of a re-identification system’s performance.

\subsection{Comparative Performance Analysis}
\input{datatables}
In Table \ref{tab:prcc_vc_table} and Table \ref{tab:celebrity_table}, we show the comparison between our method and other methods on PRCC, VC-Clothes, and Celeb-reID datasets.

In Table \ref{tab:prcc_vc_table}, the data on PRCC dataset shows that all methods perform well on the Same-Clothes subset of PRCC dataset, achieving good performance in both Rank-1 and mAP metrics. However, when identifying only Cloth-Changing images, multiple methods show significant decrease in Rank-1. But the method proposed in this paper performs excellently on the Cloth-Changing dataset, with the widely recognized Rank-1 metric decreasing by only 33.2\% in the clothing-change dataset, and at 66.4\% exceeds all comparison methods, while only slightly behind the GCA method in mAP. Looking at the VC-Clothes dataset, our method surpasses other comparison methods in both Rank-1 and mAP.
\input{celebdataset}
\input{ablation_study}
Table \ref{tab:celebrity_table} shows the comparison results between various methods and our method on the Celeb-reID dataset. Our TSDW method outperforms other comparison methods in both Rank-1 and mAP performance, and our model remains the best performing on Celeb-reID.

Considering the performance across PRCC, VC-Clothes and Celeb-reID datasets, our method achieves satisfactory results in clothing-change datasets. Our method is superior to other methods, which further proves the advancement of our proposed TSDW.

\subsection{Ablation Study Analysis}

Table \ref{tab:Ablation Study Table} shows the results of ablation experiments conducted on Celeb-reID, PRCC, and VC-Clothes datasets. The first three rows display the individual performance of each stream, the middle three rows show the complementary performance of simple pairwise combinations, the second-to-last row shows the results of simply combining all three streams, and the final row presents the results of three-stream fusion using dynamic weighting. Evidently, in the first two datasets, the Face stream consistently outperforms the other two streams, as faces typically provide more invariant features. However, in the VC-Clothes CC dataset, the Face stream performs at least 50\% worse than the other two streams, possibly because this dataset is derived from games rather than the real world, resulting in less prominent facial features, thus requiring compensation from the other two streams.

In the simple pairwise combinations, we can observe that the Rank-1 performance on Celeb-reID and PRCC is either superior to or comparable with individual streams. However, on VC-Clothes, the Rank-1 performance is not always better than individual streams. Clearly, this occurs because after the Face stream is compromised or less prominent, the fusion lacks a more intelligent method and relies on simple fusion. Although combinations containing the Face stream show improved performance, other streams that should have performed better are negatively affected.

When combining all three streams, the performance on Celeb-reID and VC-Clothes exceeds previous configurations, but the performance on PRCC does not surpass that of Face+Global. This indicates that the Head-Limb stream has become the model's bottleneck, and its limitations prevent the three streams from achieving their theoretical optimal performance.

The final Face+Head-Limb+Global+DWT represents our proposed model, which uses dynamic weighting three-way decision to enable complementary advantages across all streams. This demonstrates that our Tri-Stream Dynamic Weight Network method can significantly enhance model accuracy in clothing-changing person re-identification (CC-ReID) and, through its unique dynamic weighting approach, offsets the problems caused by the loss or lack of prominence of single features, effectively improving model robustness.

\section{CONCLUSION}
In this paper, to apply ReID technology in more general scenarios, we have proposed a Tri-Stream Dynamic Weight Network (TSDW) building upon existing Clothing Change Person Re-Identification (CC-ReID). By exploring applicable data for facial stream, head-and-limb stream, and clothing-invariant global stream across diverse scenarios, we defined a dynamic weighted decision confidence network that uses all clothing-invariant information available in images as the basis for decision-making. This approach significantly improves both prediction accuracy and model robustness to low-quality data. Experiments on multiple datasets (e.g., PRCC, Celeb-reID, VC-Clothes) demonstrate that TSDW substantially outperforms existing methods that rely on only one feature for judgment. Through extensive experimentation, we have validated the advanced nature and superior robustness of our proposed method.

\printbibliography 
\end{multicols}
\end{document}

%% file: pseudocode.tex
\begin{algorithm}[H]
\caption{Dynamic Weighted Three-Decisions Module}
\SetAlgoLined

\KwIn{Query features $q = \{q_f, q_l, q_g\}$, Gallery features $g = \{g_f, g_l, g_g\}$}
\KwOut{Final distance $D_{final}$}

\tcp{First Layer: Face reliability evaluation}
$C_f \gets \sigma(confidence_f(q_f, g_f))$\;

\uIf{$C_f > \alpha_f$}{
    $\mathbf{w} \gets [1, 0, 0]$ \tcp*{Use only face features}
}
\uElseIf{$\beta_f \leq C_f \leq \alpha_f$}{
    \tcp{Second Layer: Face partially reliable}
    $C_{l-g_1} \gets \sigma(confidence_{l-g_1}([q_l; q_g], [g_l; g_g]))$\;
    
    \uIf{$C_{l-g_1} > \alpha_{l-g_1}$}{
        $\mathbf{w} \gets Gating_{f-g}(q_f, q_g, g_f, g_g)$ \tcp*{Face-global fusion}
    }
    \uElseIf{$C_{l-g_1} < \beta_{l-g_1}$}{
        $\mathbf{w} \gets Gating_{f-l}(q_f, q_l, g_f, g_l)$ \tcp*{Face-limb fusion}
    }
    \Else{
        $\mathbf{w} \gets Gating_{all}(q_f, q_l, q_g, g_f, g_l, g_g)$ \tcp*{All features fusion}
    }
}
\Else{
    \tcp{Third Layer: Face unavailable/unreliable}
    $C_{l-g_2} \gets \sigma(confidence_{l-g_2}([q_l; q_g], [g_l; g_g]))$\;
    
    \uIf{$C_{l-g_2} > \alpha_{l-g_2}$}{
        $\mathbf{w} \gets [0, 0, 1]$ \tcp*{Use only global features}
    }
    \uElseIf{$C_{l-g_2} < \beta_{l-g_2}$}{
        $\mathbf{w} \gets [0, 1, 0]$ \tcp*{Use only limb features}
    }
    \Else{
        $\mathbf{w} \gets Gating_{l-g}(q_l, q_g, g_l, g_g)$ \tcp*{Limb-global fusion}
    }
}

\tcp{Compute feature distances and perform weighted fusion}
$d_f \gets 1 - \frac{q_f \cdot g_f}{||q_f|| \cdot ||g_f||}$\;
$d_l \gets 1 - \frac{q_l \cdot g_l}{||q_l|| \cdot ||g_l||}$\;
$d_g \gets 1 - \frac{q_g \cdot g_g}{||q_g|| \cdot ||g_g||}$\;

\tcp{Assuming $\mathbf{w} = [w_f, w_l, w_g]$ are the dynamically allocated weights}
$D_{final} \gets \mathbf{w}[1] \cdot d_f + \mathbf{w}[2] \cdot d_l + \mathbf{w}[3] \cdot d_g$\;

\Return $D_{final}$
\end{algorithm}

%% file: datatables.tex
\begin{table*}
    \centering
    \begin{tabular}{|c|c|c|c|c|c|c|c|c|} \hline  
         \multirow{3}*{Methods}&  \multicolumn{4}{c|}{PRCC}&  \multicolumn{4}{c|}{VC-Clothes}\\ \cline{2-9}  
         &  \multicolumn{2}{c|}{Same-Clothes}&  \multicolumn{2}{c|}{Cloth-Changing}&  \multicolumn{2}{c|}{Same-Clothes}&  \multicolumn{2}{c|}{Cloth-Changing}\\ \cline{2-9}  
         &  Rank-1&  mAP&  Rank-1&  mAP&  Rank-1&  mAP&  Rank-1&  mAP\\ 
         \hline  
         PCB (2018)\cite{sun2018modelspersonretrievalrefined}&  86.9&  83.6&  22.9&  24.7&  72.3&  73.9&  53.9&  55.6\\ 
         \hline  
         HACNN(2018)\cite{li2018harmoniousattentionnetworkperson}&  82.4&  84.7&  21.8&  23.2&  68.6&  69.7&  49.6&  50.1\\ 
         \hline  
         MGN(2018)\cite{Wang_2018}& 89.8& 87.4& 25.9& 35.9& 74.3& 75.2& 55.0& 57.3\\ 
         \hline  
         TransReID(2021)\cite{he2021transreidtransformerbasedobjectreidentification}& 93.1& 94.0& 40.1& 43.6& 79.8& 80.3& 73.1& 74.9\\ 
         \hline  
         SE+CESD(2020)\cite{liu2022long}& 91.8& 90.6& 37.6& 38.7& 85.2& 79.1& 69.5& 65.5\\ \hline  
         3DSL(2021)\cite{9578604,Wang_2023}& 98.7\textsuperscript{\#}& 95.0\textsuperscript{\#}& 51.3& 49.8\textsuperscript{\#}& 92.5\textsuperscript{\#}& 79.7\textsuperscript{\#}& 79.9& 81.2\\ \hline  
         UCAD(2022)\cite{ijcai2022p212}& 96.5& 95.9& 45.3& 45.2& 92.6& 81.1& 82.4& 73.8\\ \hline
         MVSE(2022)\cite{gao2021multigranularvisualsemanticembeddingclothchanging,Wang_2023}& 98.7\textsuperscript{\#}& 98.3\textsuperscript{\#}& 47.4& 52.5& 86.1\textsuperscript{\#}& 79.5\textsuperscript{\#}& 79.4\textsuperscript{\#}&79.1\textsuperscript{\#}\\\hline
         M2NET(2022)\cite{liu2022long}& 99.5& 99.1& 59.3& 57.7& /& /& /&/\\\hline
         CAL(2022)\cite{Gu_2022}& 100& 99.8& 55.2& 55.8& /& /& /&/\\\hline
         AFL(2023)\cite{10269038}& 100& 99.7& 57.4& 56.5& 93.9& 88.3& 82.5&83.0\\\hline
         AIM(2023)\cite{GoodisBad}& \textbf{100}& \textbf{99.9}& 57.9& 58.3& /& /& /&/ \\\hline
         GCA(2024)\cite{Ding_2024}& 99.3& 94.3& 64.8& \textbf{61.3}& 93.1& 92.8& 83.7&82.7\\\hline
         \textbf{TSDW(ours)}& 99.6& 96.6& \textbf{66.4}& 58.8& \textbf{94.5}& \textbf{94.7}& \textbf{88.0}&\textbf{87.1}\\\hline
    \end{tabular}
    \caption{\textbf{Comparison on PRCC and VC-Clothes Datasets.} \\ 
     The \textsuperscript{\#} at the end indicates that this specific data point is based on reproduction experiments by Wang et al.~\cite{Wang_2023}.}
    \label{tab:prcc_vc_table}
\end{table*}

%% file: celebdataset.tex
\begin{table*}
    \centering
    \begin{tabular}{|c|c|c|} \hline  
         Methods & Rank-1 & mAP \\ 
         \hline  
         PCB(2018)~\cite{sun2018modelspersonretrievalrefined} & 45.1 & 8.7 \\ 
         \hline
        ReIDCaps+ (2020)~\cite{8873614} & 63.0 & 15.8 \\
         \hline
         CASE-Net (2021)~\cite{li2021learning} & 66.4 & 18.2 \\
         \hline
        LightMBN (2021)~\cite{herzog2021lightweight} & 59.2 & 15.2 \\
         \hline
         AFD-Net (2021)~\cite{xu2021adversarial} & 52.1 & 10.6 \\
         \hline
         IS-GAN\(\mathit{}_{\text{Best}}\)(2022)\cite{eom2021disentangled}& 54.9&14.5\\\hline
         SirNet (2022)~\cite{sirNet2022} & 56.0 & 14.2 \\
         \hline
         IRANet (2022)~\cite{shi2022iranet} & 64.1 & 19.0 \\
         \hline
         3DInvarReID (2023)~\cite{liu2023learningclothingposeinvariant} & 65.5 & 18.4 \\
         \hline
         VersReID (2024)~\cite{zheng2024versatileframeworkmultisceneperson} & 61.7 & 18.7 \\
         \hline
         CSSC (2024)~\cite{wang2024contentsalientsemanticscollaboration} & 64.5 & 17.3 \\
         \hline
         FIRe (2024)~\cite{Wang_2024} & 64.0 & 18.2 \\
         \hline
         \textbf{TSDW (ours)} & \textbf{67.2} & \textbf{19.3} \\
         \hline
    \end{tabular}
    \caption{\textbf{Comparison on Celeb-reID Dataset}}
    \label{tab:celebrity_table}
\end{table*}

%% file: ablation_study.tex
\begin{table*}
    \centering
    \begin{tabular}{|c|c|c|c|c|c|c|c|c|c|c|c|} \hline  
         \multicolumn{4}{|c|}{Method}& \multicolumn{2}{|c|}{Celeb-reID} & \multicolumn{4}{|c|}{PRCC} & \multicolumn{2}{|c|}{VC-Clothes CC} \\ 
         \hline  
        \multirow{2}{*}{Face} & \multirow{2}{*}{\begin{tabular}{c}Head-\\Limb\end{tabular}} & \multirow{2}{*}{Global} & \multirow{2}{*}{DWT} & \multirow{2}{*}{Rank-1} & \multirow{2}{*}{mAP} & \multicolumn{2}{c|}{Same-Clothes} & \multicolumn{2}{c|}{Cloth-Changing} & \multirow{2}{*}{Rank-1} & \multirow{2}{*}{mAP} \\ 
        \cline{7-10} 
         & & &  & & & Rank-1 & mAP & Rank-1 & mAP & & \\ \hline  
         
         \checkmark & &  & & 54.3 & 11.6 & 71.9 & 44.5 & 61.4 & 40.1 & 36.1 & 35.0 \\ \hline  
         
         & \checkmark &  & & 52.3 & 9.8 & 74.3 & 57.0 & 49.4 & 39.2 & 73.5 & 70.1 \\ \hline  
         
         & & \checkmark  & & 53.5 & 9.1 & 100 & 99.7 & 51.3 & 54.0 & 77.8 & 79.3 \\ \hline 
 \checkmark& \checkmark& & & 59.8& 14.1& 90.3& 77.2& 62.9& 51.0& 70.0&62.4\\ \hline 
 \checkmark& & \checkmark& & 58.6& 13.7& 99.7& 97.5& 65.9& 58.6& 79.8&77.1\\ \hline 
 & \checkmark& \checkmark& & 58.7& 12.3& \textbf{99.9}& 98.2& 57.6& 55.8& 84.1&84.3\\ \hline 
 \checkmark& \checkmark& \checkmark& & 62.9& 14.7& 99.8& \textbf{98.5}& 63.4& \textbf{59.5}& 85.5&84.6\\ \hline 
         
         \checkmark& \checkmark&  \checkmark& \checkmark& \textbf{67.2} & \textbf{19.3} & 99.6 & 96.6 & \textbf{66.4} & 58.8 & \textbf{88.0} & \textbf{87.1} \\ \hline 
         
    \end{tabular}
    \caption{\textbf{Ablation Study Table}}
    \label{tab:Ablation Study Table}
\end{table*}